\documentclass[runningheads]{llncs}
\usepackage[T1]{fontenc}
\usepackage{graphicx,verbatim,amsmath,multirow,amssymb}
\usepackage{siunitx}
\usepackage{booktabs}
\usepackage{amssymb}

\begin{document}
\title{Stitch-Inferencer: Enhance Endoscopic Video Segmentation and Tracking via Panoramic Reconstruction}
\titlerunning{Stitch-Inferencer}
%

\author{Shunsuke Kikuchi\inst{1}\orcidID{0009-0008-3353-657X} \and
        Atsushi Kouno\inst{1}\orcidID{0000-0002-6365-7638} \and
        Hiroki Matsuzaki\inst{1}
        }
\authorrunning{S. Kikuchi et al.}
\institute{Jmees Inc, Kashiwa-city, Chiba, Japan \\
    \email{engineer@jmees-inc.com}}

\maketitle              
\begin{abstract}
Surgical video understanding is fundamental to navigation systems. Endoscopic perception is often hindered by a limited field-of-view and frequent instrument occlusions, making spatio-temporal context essential for robust inference. These challenges have motivated video models that aggregate information across frames. However, existing video models typically store past observations implicitly in learned feature representations, often requiring task-specific video training, substantial annotated data, and increased computational cost.
We propose \textbf{Stitch-Inferencer}, a real-time, model-agnostic inference framework that replaces implicit feature memory with an explicit image-space panoramic canvas. By stitching valid observations across frames, Stitch-Inferencer preserves previously observed pixels in an online, instrument-free view, expanding the effective field-of-view and providing direct access to regions that are temporarily occluded or absent from the current frame. Downstream segmentation or tracking models are applied to a compact region of interest on the panorama, and their predictions are reprojected to the current frame, enabling existing models to exploit long-range context without retraining. Experiments on anatomy segmentation and point/box tracking demonstrate consistent improvements across diverse baselines while preserving real-time throughput. The stitching module alone runs at over 60 FPS, providing a practical inference-time solution to enhance endoscopic perception in computationally constrained intraoperative environments. Source code will be made publicly available.
\end{abstract}
\section{Introduction}
\begin{figure}[!ht]
    \centering
    \includegraphics[width=\linewidth]{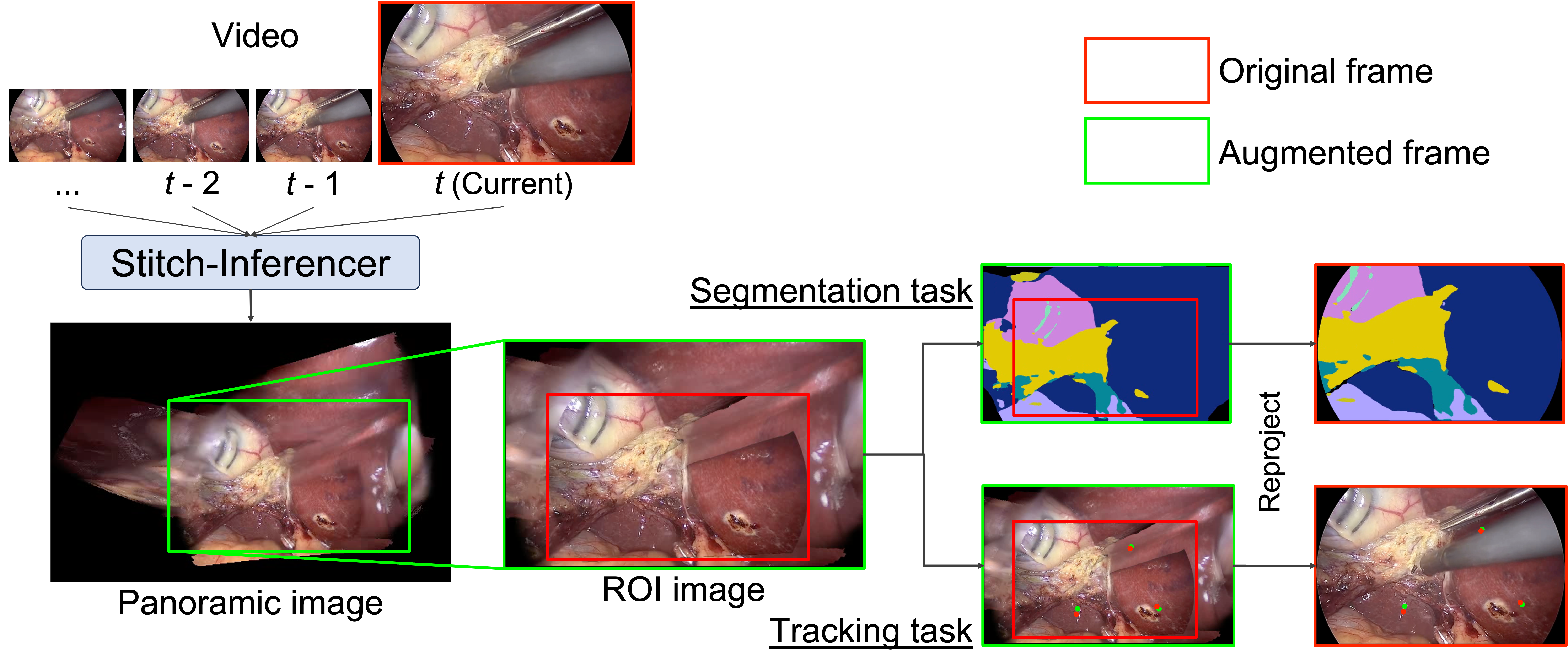}
    \caption{Graphical abstract. Stitch-Inferencer creates an instrument-free panoramic view and improves downstream segmentation and tracking at inference time without additional training; green and red boxes indicate the panorama context and the current-frame region with reprojection, respectively.}
    \label{fig:graphical-abstract}
\end{figure}

Endoscopic perception in intraoperative settings is often characterized by partial observability and tight latency requirements, especially for surgical navigation and assistance systems that must update reliably in real time. Clinically relevant anatomy may lie outside the narrow field-of-view, be temporarily occluded by instruments, or only become interpretable when compared against earlier observations. While per-frame recognition, segmentation, and tracking have advanced rapidly, relying on the current frame alone remains brittle in realistic procedures, motivating the need for long-range context in surgical video understanding.

A dominant approach to incorporating such context is to learn temporal memory through recurrent architectures, temporal attention, spatio-temporal convolution, or recent sequence models~\cite{tmanet,sptcn,Park2025VideoMamba}. Although effective, these methods typically store past observations implicitly in latent feature representations. As a result, previously visible anatomical regions are not explicitly preserved once they leave the current field-of-view or become occluded; the model must instead rely on compressed feature states to recover or infer the missing context. This implicit form of memory can be powerful when trained with sufficient data, but it also increases architectural complexity and often requires task-specific training to exploit temporal information effectively.

We revisit a simple and comparatively underexplored alternative: maintaining memory directly in the image space. Recent advances in image matching make robust 2D alignment increasingly practical \cite{aliked,lightglue}. In a large class of laparoscopic procedures, camera motion is moderate over short windows and the view is often approximately perpendicular to the anatomical surface. This makes it possible to accumulate previously observed anatomy into an online panoramic canvas. Unlike latent feature memory, such a canvas preserves observed pixels explicitly, expands the effective field-of-view, and provides direct access to regions that may be temporarily hidden by instruments or absent from the current frame. In this sense, the panorama acts as an explicit spatial memory for downstream inference.

Based on this idea, we propose \textbf{Stitch-Inferencer}, a real-time, model-agnostic framework that constructs an online, instrument-free panoramic view and performs inference on the panorama with projection back to the current frame. At each time step, invalid regions such as image borders, instruments, and trocar ports are masked, valid observations are aligned and composited into a global canvas, and a compact region-of-interest around the current frame is extracted for downstream inference. Existing segmentation or tracking models can then be applied directly to this augmented view without retraining, and their predictions are reprojected to the original frame coordinates. This design upgrades single-frame models with temporally accumulated context at inference time while preserving online operation.

We evaluate Stitch-Inferencer on anatomy segmentation and point/box tracking and observe consistent improvements across diverse baselines while preserving real-time inference speed. Our contributions are: (1) an online method for constructing an interpretable explicit image-space spatial memory from endoscopic video, producing an instrument-free panoramic view robust to common surgical artifacts; (2) a model-agnostic projection/inverse-projection inference module that enhances existing models with long-range context at inference time, without retraining; and (3) empirical gains on segmentation and tracking under real-time constraints.

\section{Methods}
\label{sec:methods}

\begin{figure}[!t]
    \centering
    \includegraphics[width=\linewidth]{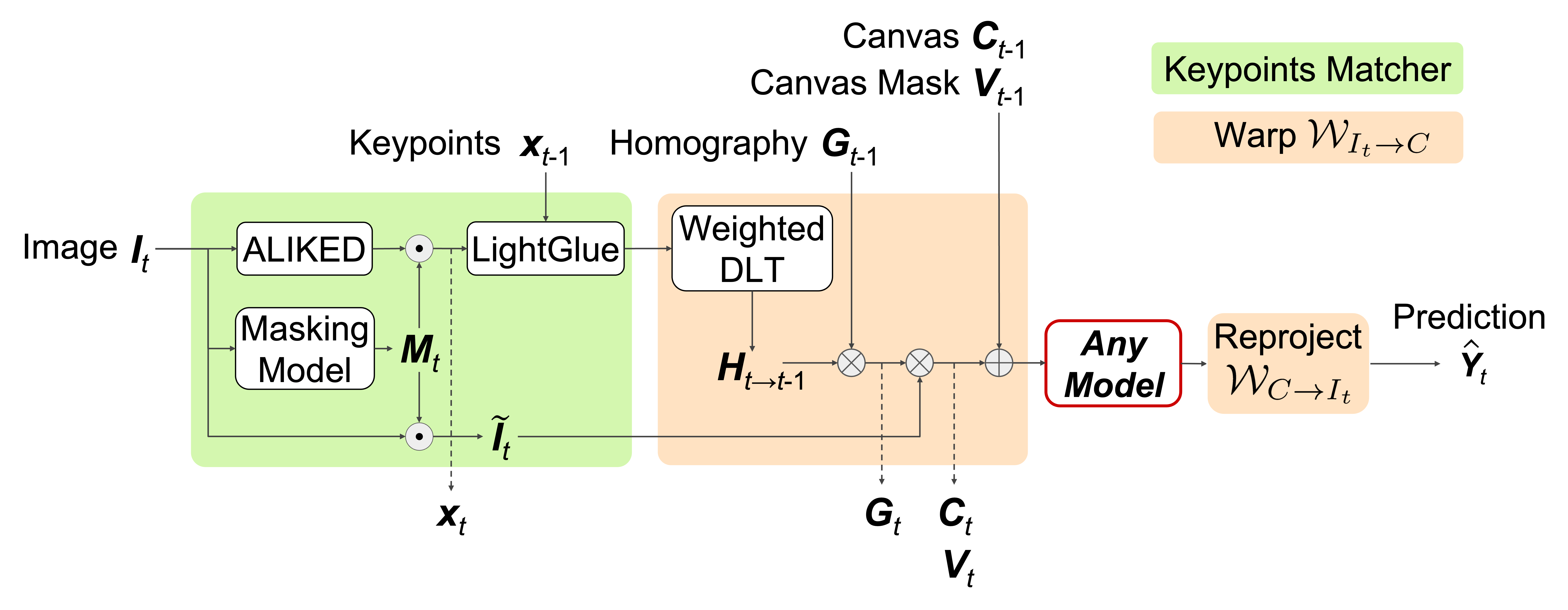}
    \caption{Overview of the Stitch-Inferencer pipeline. Green: keypoint matching and homography estimation. Orange: warping, canvas update, and reprojection.}
    \label{fig:method-overview}
\end{figure}

Stitch-Inferencer maintains an online instrument-free panoramic view and enables inference on the panorama with projection back to the current frame. The module is a lightweight state machine defined by a canvas image $C_t$, a canvas valid mask $V_t$, and a cumulative warp $G_t$. Given frame $I_t\!\in\!\mathbb{R}^{h\times w\times 3}$, we allocate a canvas of size $3h\times 3w$ (optionally scaled). Let $T$ translate the frame origin to the canvas center. We denote image warping by $\mathcal{W}(I;G)$ (bilinear for RGB, nearest for masks).
At each time step, the pipeline in Fig.~\ref{fig:method-overview} proceeds in five stages:
(i) mask invalid regions;
(ii) estimate the inter-frame homography from matched keypoints;
(iii) update the cumulative warp and derive the frame--canvas mapping;
(iv) warp and composite the frame into the canvas;
(v) run a downstream model on the canvas and reproject it to the current frame.

\textbf{Invalid-region mask.}
We form an invalid mask $M_t = M_t^{\text{border}} \cup M_t^{\text{tool}} \cup M_t^{\text{port}}$.
$M_t^{\text{border}}$ is obtained by 8-bit grayscale thresholding of extremely dark pixels ($\le 15$), fitting an ellipse to the visible circular field-of-view, and marking pixels outside the ellipse as invalid.
$M_t^{\text{tool}}$ and $M_t^{\text{port}}$ are produced by ConvNeXt-tiny U-Net models~\cite{unet,convnext} trained on multi-dataset surgical tool annotations~\cite{cholecinstanceseg,segcol,surgtoolloc,endoscapes1,phakir,sarrarp50} and on Cholec80-port~\cite{cholec80port} (with label alignment on m2caiSeg~\cite{m2caiseg} and GynSurg~\cite{gynsurg}) using the same image-training protocol as anatomy segmentation baselines (input $512{\times}512$, AdamW~\cite{adamw}, learning rate $5{\times}10^{-5}$, affine, blur and photometric augmentations).
We define a masked frame $\tilde I_t = I_t \odot (1 - M_t)$.

\textbf{Matching and inter-frame homography.}
We detect and describe keypoints with ALIKED~\cite{aliked} and match with LightGlue~\cite{lightglue}.
Keypoints falling inside $M_t$ are discarded.
For tracking experiments only, we additionally apply flow-based keypoint filtering (particularly important for tracking, which is more sensitive to pixel-level misalignment). We compute dense optical flow using NeuFlowv2~\cite{neuflowv2} and discard keypoints whose flow magnitude is a $2\sigma$ outlier after standardization: letting $m(\mathbf{x}_t)$ denote the flow-magnitude map, we compute its mean $\mu_t$ and standard deviation $\sigma_t$ over valid pixels and discard keypoints at locations $\mathbf{x}_{t,i}$ satisfying $|m(\mathbf{x}_{t,i})-\mu_t|>2\sigma_t$.

From the remaining correspondences $\{(\mathbf{x}_{t,i},\mathbf{x}_{t-1,i},w_i)\}$ where match confidence weight $w_i$ is given by LightGlue, we estimate $H_{t\to t-1}$ using weighted Direct Linear Transform (DLT). If fewer than four matches remain, we re-initialize $(C_t,V_t,G_t)$ from $I_t$.

\textbf{Cumulative warp and frame-to-canvas mapping.}
We update the cumulative warp $G_t = H_{t\to 0} = H_{t\to t-1}\,G_{t-1}$ with $G_0=\mathbf{I}_3$.
The current frame-to-canvas warp is
$\mathcal{W}_{I_t \to C}(I):= W(I; S\,T\,G_t) $, and its inverse mapping is $\mathcal{W}_{C \to I_t} := \mathcal{W}^{-1}_{I_t \to C}$,
where $S$ is an optional scale matrix and $T$ centers the current frame on the canvas.
If the implied update is extreme (shear $>15^\circ$, rotation $>15^\circ$, or scale $>2.0$), we re-anchor the panorama by re-centering the canvas on the current frame, set $G_t = \mathbf{I}_3$, and restart accumulation from this frame.

\textbf{Compositing and Update States.}
We compute a Laplacian-variance blur score $b_t$ on the current frame and skip the storage update when $b_t<\tau_{\text{blur}}$ ($\tau_{\text{blur}}=30$), i.e., we keep $(C_t,V_t)=(C_{t-1},V_{t-1})$ for the stored panorama.
If $b_t \ge \tau_{\text{blur}}$, we update the canvas by warping the masked frame and alpha-blending only on valid pixels:
\begin{align}
\tilde I_t^C &= \mathcal{W}_{I_t\to C}(\tilde I_t), \quad  V_t^{\text{new}} = \mathcal{W}_{I_t\to C}(1-M_t), \quad V_t = V_{t-1} \cup V_t^{\text{new}} \\
C_t &= \alpha_t \odot C_{t-1} + (1-\alpha_t)\odot \tilde I_t^C \quad \text{(applied only where } V_t^{\text{new}}=1\text{)}.
\end{align}
where $\alpha_t^C=\mathcal{W}(\alpha_t^I;H_{t\to C})$ is a seam-aware blending weight. 
To smooth blending seams, we compute a spatial alpha-blending weight $\alpha_t$ based on a distance transform from the combined boundary of the invalid mask $M_t$ and the frame edges. $\alpha_t$ decays linearly up to $r=201$ pixels from the boundary (computed via iterative morphological dilation at $1/8$ scale).

\textbf{Inference ROI and reprojection.}
For inference, we form an inference canvas by compositing the current masked frame onto the latest stored $C_t$, then crop a region of interest (ROI) that (i) contains the entire current frame content in canvas coordinates and (ii) lies within the valid stitched region $V_t$. The ROI is resized to the model input and we obtain predictions in ROI coordinates. 
For segmentation, we paste the ROI prediction back to the original ROI location on the canvas (zero elsewhere) to obtain $\hat{Y}_t^C$, then reproject to the current frame by $\hat{Y}_t=\mathcal{W}_{C\to I_t}(\hat{Y}_t^C)$. For tracking, we map points by $\hat{\mathbf{y}}_t \sim (S\,T\,G_t)^{-1} \hat{\mathbf{y}}_t^C$, followed by homogeneous normalization.

\section{Experiments and Results}
\label{sec:experiments}
We evaluated Stitch-Inferencer on two downstream tasks: anatomy segmentation and tracking. For both tasks, we compared original-frame inference with the same models augmented by Stitch-Inferencer, without any fine-tuning. For anatomy segmentation, we additionally compared against temporal and specialized video segmentation baselines. We also provide qualitative examples and ablate the inference-view construction to analyze how panoramic context should be provided to downstream models. Segmentation training and tracking experiments were conducted on an NVIDIA Quadro RTX 8000 with an AMD EPYC 7702P CPU, and segmentation inference speed was measured on an NVIDIA GeForce RTX 4090 with an Intel Core i9-10980XE CPU.

\subsection{Anatomy segmentation}
\label{sec:experiments:seg}

\begin{table}[!t]
\caption{Video segmentation results with and without panorama inference. ``Stitch'' indicates inference on the stitched panorama with projection back to the current frame.}
\label{tab:segmentation-main}
\centering
{\fontsize{8}{9}\selectfont
\setlength{\aboverulesep}{1.5pt}
\setlength{\belowrulesep}{1.5pt}
\begin{tabular}{
l l c
S[table-format=1.4]
S[table-format=1.4]
S[table-format=3.2]
S[table-format=3.3]
}
\toprule
Method & Encoder & Stitch &
{Dice$\uparrow$} & {peri-Dice$\uparrow$} & {FPS$\uparrow$} & {p95 latency$\downarrow$} \\
\midrule
\multirow{2}{*}{UNet++~\cite{unetplusplus}} & \multirow{2}{*}{ConvNeXt-B} &  & 0.6884 & 0.7159 & 205.89 & 4.921 \\
 &  & {\checkmark} & 0.7436 & 0.7541 & 49.35 & 20.598 \\
\midrule
\multirow{2}{*}{UPerNet~\cite{upernet}} & \multirow{2}{*}{ConvNeXt-B} &  & 0.7100 & 0.7489 & 212.17 & 4.780 \\
 &  & {\checkmark} & 0.7195 & 0.7623 & 47.10 & 21.522 \\
\midrule
\multirow{2}{*}{SegFormer~\cite{segformer}} & \multirow{2}{*}{MiT-B5} &  & 0.7128 & 0.7788 & 104.89 & 10.136 \\
 &  & {\checkmark} & 0.6995 & 0.8008 & 45.03 & 22.512 \\
\midrule
\multirow{2}{*}{DeepLabV3+~\cite{deeplabv3}} & \multirow{2}{*}{ResNet-101} &  & 0.6813 & 0.7566 & 159.32 & 6.277 \\
 &  & {\checkmark} & 0.7044 & 0.7525 & 47.72 & 21.272 \\
\midrule
\multirow{2}{*}{HRNet-seg~\cite{hrnet}} & \multirow{2}{*}{HRNet-32} &  & 0.6441 & 0.7315 & 119.74 & 8.816 \\
 &  & {\checkmark} & 0.6508 & 0.7362 & 51.72 & 19.651 \\
\specialrule{0.4pt}{0pt}{0pt}
\specialrule{0.4pt}{0.7pt}{1.72pt}
SP-TCN~\cite{sptcn} & HRNet-32 & & 0.6016 & 0.6978 & 4.31 & 231.944 \\
TMANet~\cite{tmanet} & HRNet-32 &  & 0.5050 & 0.3257 & 9.47 & 105.556 \\
DTERN~\cite{dtern} & MiT-B5 &  & 0.7006 & 0.6854 & 11.26 & 98.782 \\
SALI~\cite{sali} & PVTv2~\cite{pvtv2} & & 0.6744 & 0.6976 & 14.65 & 70.394 \\
UPerNet & VideoMamba~\cite{Park2025VideoMamba} &  & 0.7130 & 0.7517 & 13.44 & 74.853 \\
\specialrule{0.4pt}{0pt}{0pt}
\specialrule{0.4pt}{0.7pt}{1.72pt}
MedicalSAM3~\cite{medicalsam3} & {(box prompt)} &  & 0.5933 & 0.7256 & 2.55 & 393.933 \\
SAM3~\cite{sam3} & {(point prompt)} & & 0.3398 & 0.3257 & 2.28 & 442.724 \\
\bottomrule
\end{tabular}}
\end{table}

\begin{figure}[!b]
    \centering
    \includegraphics[width=\linewidth]{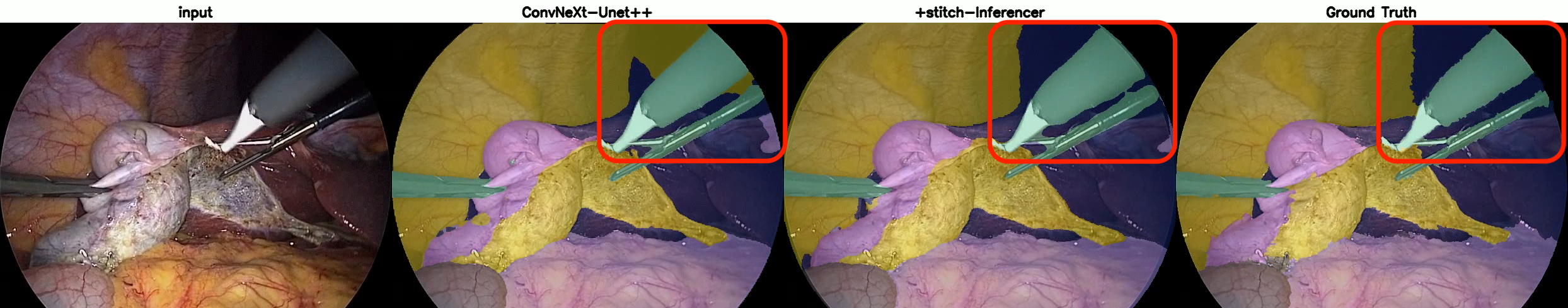}
    \caption{Qualitative anatomy segmentation example on CholecSeg8k with ConvNeXt-B UNet++. From left to right: input frame, original inference, Stitch-Inferencer inference, and ground truth.
Red boxes highlight a peripheral region where Stitch-Inferencer better recovers the ground-truth segmentation by leveraging panoramic spatial context.}
    \label{fig:pred_seg}
\end{figure}

We evaluated anatomy segmentation on the CholecSeg8k~\cite{cholecseg8k} dataset. Since CholecSeg8k frame indices were not perfectly aligned with the original Cholec80 videos, we first matched frames to Cholec80 timestamps to enable evaluation under video inference. We used 5-fold cross-validation: rare-class clips (videos 01, 09, 12) were split with stratified $K$-fold at the clip level, and all others used stratified group $K$-fold grouped by Cholec80 video to minimize leakage.

We tested Stitch-Inferencer on five single-frame baselines (Table~\ref{tab:segmentation-main}) under a unified training protocol, and followed the original settings for additional temporal/specialized baselines. We reported macro Dice excluding background and tool classes (grasper, L-hook electrode), and additionally peri-Dice (outside a central ellipse, radius $>2/3$). Speed was measured over 10{,}000 frames using FPS and p95 latency.

As shown in Table~\ref{tab:segmentation-main}, panorama inference improved Dice for all evaluated single-frame baselines except SegFormer, without retraining or architectural changes. The mean Dice gain over sequences was $+0.0553\pm0.0174$ (UNet++), $+0.0096\pm0.0051$ (UPerNet), $-0.0133\pm0.0127$ (SegFormer), $+0.0231\pm0.0071$ (DeepLabV3+), and $+0.0067\pm0.0046$ (HRNet-seg). Gains were also generally consistent on peri-Dice, indicating improved robustness under limited field-of-view. The stitched variants still ran at practical real-time throughput (45--52 FPS). Figure~\ref{fig:pred_seg} provides qualitative examples of how the panoramic spatial memory changes downstream inference in dark region at screen edge. Overall, Stitch-Inferencer enhanced diverse segmentation models at inference time while maintaining practical throughput.

In addition to downstream inference speed, the stitching module alone (TensorRT) ran at 63.35 FPS with 16.049 ms p95 latency on the full Cholec80 dataset. Canvas re-initialization due to low matching occurred 4{,}207 times over 4{,}612{,}530 frames (0.092\%), mainly in extra-corporeal scenes or when instruments occupied most of the view. The masks used for invalid-region removal achieved Dice scores of 0.86 for tool segmentation on EndoVis18~\cite{endovis18} and 0.77 for port segmentation on the Cholec80-port test set.

\subsection{Tracking}
\label{sec:experiments:track}

We evaluated on STIR24~\cite{STIR24} (point tracking) and SurgT~\cite{surgt} (3D box tracking). For STIR24, we reported threshold accuracy average ($\delta$-avg) at $\delta\in\{4,8,16,32,64\}$ pixels and mean distance error (pixels). For SurgT, we tracked the 2D box center in left/right views, reconstructed 3D trajectories by stereo pairing, kept box size fixed to its initial value, and reported Expected Average Overlap (EAO), mean 2D error (px), and mean 3D error (mm). All trackers used public weights with no fine-tuning, and we kept the original video resolution.

\begin{table}[t]
\caption{Comparison of tracking methods on STIR and SurgT datasets with and without Stitch-Inferencer.}
\label{tab:tracking-results}
\centering
{\fontsize{8}{9}\selectfont
\setlength{\aboverulesep}{1.5pt}
\setlength{\belowrulesep}{1.5pt}
\begin{tabular}{l c c c c c c}
\toprule
& & \multicolumn{2}{c}{STIR24} & \multicolumn{3}{c}{SurgT} \\
\cmidrule(lr){3-4}\cmidrule(lr){5-7}
Method & Stitch & {$\delta$-avg$\uparrow$} & {mean distance$\downarrow$} & {EAO$\uparrow$} & {Err2D$\downarrow$} & {Err3D$\downarrow$} \\
\midrule
\multirow{2}{*}{CSRT~\cite{csrt}} &  & 0.62246 & 44.229 & 0.552 & 6.0 $\pm$ 2.7 & 3.51 $\pm$ 3.8 \\
 & {\checkmark} & 0.65908 & 25.807 & 0.579 & 5.7 $\pm$ 2.7 & 3.36 $\pm$ 2.5 \\
\midrule
\multirow{2}{*}{Cotracker3~\cite{cotracker3}} &  & 0.70087 & 19.100 & 0.453 & 13.0 $\pm$ 11.0 & 3.13 $\pm$ 2.8 \\
 & {\checkmark} & 0.71634 & 16.029 & 0.602 & 12.7 $\pm$ 11.1 & 2.79 $\pm$ 2.4 \\
\midrule
\multirow{2}{*}{Litetracker~\cite{litetracker}} &  & 0.76298 & 17.150 & 0.585 & 5.1 $\pm$ 1.5 & 1.55 $\pm$ 1.1 \\
 & {\checkmark} & 0.76215 & 14.906 & 0.685 & 5.2 $\pm$ 1.7 & 1.83 $\pm$ 1.3 \\
\midrule
\multirow{2}{*}{MFT~\cite{mft}} &  & 0.77783 & 15.716 & 0.715 & 2.2 $\pm$ 1.3 & 1.47 $\pm$ 1.7 \\
 & {\checkmark} & 0.77962 & 14.640 & 0.755 & 2.2 $\pm$ 1.2 & 1.39 $\pm$ 1.1 \\
\bottomrule
\end{tabular}}
\end{table}

Table~\ref{tab:tracking-results} shows consistent gains across all evaluated trackers on STIR24 and SurgT without tracker-specific retraining. This supported Stitch-Inferencer as a model-agnostic inference module beyond segmentation: the temporally fused instrument-free panoramic view provided a shared spatial reference that improved robustness to tool occlusions and out-of-view motion. Qualitatively, targets often lost in the original frame view could be re-associated through the panorama, further indicating that these failure modes were addressed explicitly rather than by implicit temporal memorization.

\begin{table}[b]
\caption{Ablation of inference-view ROI. Results are reported on CholecSeg8k (Dice; UNet++) and STIR24 ($\delta$-avg; MFT).}
\label{tab:ablation-results}
\centering
{\fontsize{8}{9}\selectfont
\setlength{\aboverulesep}{1.5pt}
\setlength{\belowrulesep}{1.5pt}
\begin{tabular}{l c c c c}
\toprule
Metric & Baseline & Internal & External & Entire \\
\midrule
CholecSeg8k Dice & 0.6884 & \textbf{0.7436} & 0.6802 & 0.6718 \\
STIR24 $\delta$-avg & 0.7778 & \textbf{0.7796} & 0.7789 & 0.4113 \\
\bottomrule
\end{tabular}}
\end{table}

\subsection{Ablation study}

A compact ablation on inference-view construction (Table~\ref{tab:ablation-results}). We compared the original per-frame baseline, the proposed \emph{internal} ROI, an \emph{external} ROI defined as the smallest bounding box that covered the valid stitched region ($V_t>0$), and the \emph{entire} canvas resized to the model input size. The proposed internal ROI performed best on both CholecSeg8k and STIR24, while the external and entire-canvas variants degraded performance. These results suggested that performance gains required not only panoramic context, but also an appropriately sized inference ROI: overly large views diluted relevant local detail and introduced unnecessary background, whereas the proposed ROI preserved a compact, well-aligned context around the current frame.

\section{Discussion and Conclusion}
\label{sec:discussion}

This work shows that an instrument-free panoramic view can serve as a practical, explicit spatial memory for endoscopic video understanding. Across anatomy segmentation and tracking, Stitch-Inferencer improved diverse baselines without retraining, supporting its role as a model-agnostic inference module. These results suggest that many common failure modes in endoscopic video understanding can be addressed by explicitly aggregating previously observed pixels in image space, rather than relying solely on implicit temporal memory in learned feature representations. Because the panorama is grounded in real observations rather than generative inpainting, it provides a reliable auxiliary view for surgical videos, where instrument occlusions, limited field-of-view, and intermittent visibility frequently degrade current-frame inference. Importantly, the method maintains real-time throughput while delivering measurable gains, making it compatible with intraoperative use cases such as navigation overlays and safety monitoring without requiring task-specific video-model training.

Interestingly, while CNN-based architectures consistently benefited from the expanded spatial context, SegFormer exhibited a slight performance drop. This may reflect architectural differences in context aggregation: CNN-based models rely more strongly on local feature composition and may exploit the expanded ROI, whereas SegFormer's global self-attention may be more sensitive to appearance discontinuities introduced by stitching.

Limitations primarily stem from stitching assumptions and the current evaluation ecosystem. Planar homography remains vulnerable to strong non-planar deformation and parallax; when deformable tissue moves significantly---especially exiting the field-of-view---there is no mechanism to correct the accumulated canvas, yielding unnatural panoramas or stale context. In addition, recent public benchmarks provide limited coverage of long procedures with highly dynamic camera motion, constraining evaluation under practical regimes. Similar dataset limitations with densely labeled long videos make it difficult to assess if these gains persist when stronger models are trained with substantially more data.

Several future directions remain. Fine-tuning models on panorama ROIs may further improve accuracy. Since stitching can suppress arbitrary occluders given a mask, the instrument-free panoramic view may support annotation and higher-level endoscopic video understanding tasks (e.g., phase recognition and surgical VQA). Improving robustness under deformation (e.g., confidence-aware warping/blending, uncertainty estimation, or adaptive reset policies) is an important next step for reliable deployment.



%
%
%
\pagebreak
\bibliography{citation}
\bibliographystyle{splncs04}

\end{document}